\documentclass[preprint,12pt]{elsarticle}

\usepackage{amssymb}
\usepackage{amsmath}

\usepackage{xurl}
\usepackage{hyperref}
\usepackage{cleveref}
\usepackage{kotex}
\usepackage{siunitx}
\usepackage{makecell}
\usepackage{multirow}
\usepackage{ulem}
\usepackage{adjustbox}
\usepackage{makecell}
\usepackage{threeparttable} %
\usepackage{caption} 
\usepackage{booktabs}
\usepackage{amssymb}
\usepackage{tabularx}
\usepackage[table]{xcolor}  %
\usepackage[ruled,vlined]{algorithm2e} %

\journal{Knowledge-Based Systems}

\begin{document}

\begin{frontmatter}

\title{Multi-Modal Guided Multi-Source Domain Adaptation for Object Detection}

\author[label1]{Sangin Lee\fnref{fn1}}
\ead{silee@rcv.sejong.ac.kr}

\author[label1]{Seokjun Kwon\fnref{fn1}}
\ead{sjkwon@rcv.sejong.ac.kr}

\author[label1]{Jeongmin Shin}
\ead{jmshin@rcv.sejong.ac.kr}

\author[label2]{Namil Kim}
\ead{namil.kim@naverlabs.com}

\author[label1,label3]{Yukyung Choi\corref{cor1}}
\ead{ykchoi@sejong.ac.kr}
\cortext[cor1]{Corresponding author at: Sejong University, Seoul, Republic of Korea}
\fntext[fn1]{These authors contributed equally to this work}

\affiliation[label1]{organization={Sejong University},
            city={Seoul},
            country={Republic of Korea}}

\affiliation[label2]{organization={NAVER LABS},
            city={Seongnam},
            country={Republic of Korea}}
            
\affiliation[label3]{organization={Artificial Intelligence and Robotics Institute (AIRI)},
            city={Seoul},
            country={Republic of Korea}}

\begin{abstract}
    General object detection (OD) struggles to detect objects in the target domain that differ from the training distribution. To address this, recent studies demonstrate that training from multiple source domains and explicitly processing them separately for multi-source domain adaptation (MSDA) outperforms blending them for unsupervised domain adaptation (UDA). However, existing MSDA methods learn domain-agnostic features from domain-specific RGB images while preserving domain-specific information from the domain-agnostic feature map. To address this, we propose MS-DePro: Multi-Source Detector with Depth and Prompt, composed of (1) depth-guided localization and (2) multi-modal guided prompt learning. We leverage domain-agnostic input modalities, namely depth maps and text, to encode domain-agnostic characteristics. Specifically, we utilize depth maps to generate domain-agnostic region proposals for localization and integrate multi-modal features to align learnable text embeddings for classification. MS-DePro achieves state-of-the-art performance on MSDA benchmarks, and comprehensive ablations demonstrate the effectiveness of our contributions. Our code is available on \href{https://github.com/sejong-rcv/Multi-Modal-Guided-Multi-Source-Domain-Adaptation-for-Object-Detection}{GitHub}.
\end{abstract}

% \begin{graphicalabstract}
%     \centering
%     \includegraphics[scale=0.55]{fig0_graphical_abstract.pdf}
% \end{graphicalabstract}

% \begin{highlights}
%     \item MSDA models face conflicts in learning domain-agnostic and domain-specific features.
%     \item We argue the inherent limitation of RGB images in modeling domain-agnostic features.
%     \item We leverage explicit domain-agnostic modalities, the depth map and text.
%     \item We propose novel depth-guided localization and multi-modal guided prompt learning.
%     \item Remarkable results on MSDA benchmarks validate the effectiveness of our approach.
% \end{highlights}


\end{frontmatter}

\section{Introduction}
\label{sec:intro}

    Over the past decades, object detection (OD) has witnessed remarkable progress in various challenging scenarios, including closed-set and open-world settings. However, general object detectors often struggle in target (test) domains that differ from their training source domains, as domain shifts arise from several factors such as time (day vs. night), style (real vs. synthetic), and camera artifacts (sensor, resolution). To address this issue, unsupervised domain adaptation (UDA) methods~\cite{wei2020incremental, deng2021unbiased, li2022cross, bai2024misalignment} have emerged, aiming to mitigate the domain gap between the labeled source data and the unlabeled target data. Despite existing UDA methods assuming a single-source domain, multiple datasets are available for training. Moreover, utilizing multiple sources can yield more robust and reliable performance~\cite{yao2021multi}.

    However, previous studies have shown that naively combining multiple sources and performing single-source UDA performs worse than using a single source~\cite{yao2021multi, wu2022target}. In this setting, the model struggles to learn a common latent space between multiple sources and the target domain due to the discrepancies among the source domains. To tackle this issue, multi-source domain adaptation (MSDA) for OD has been proposed~\cite{yao2021multi, wu2022target, belal2024multi, belal2024attention} to mitigate interference between multiple sources by treating each source domain separately. MSDA methods typically consist of two main components: 1) a domain-agnostic network and 2) a domain-specific network. The domain-agnostic network extracts common latent features through adversarial learning with a gradient reversal layer (GRL). As the GRL-based domain-agnostic network tends to remove high-level semantic information~\cite{DAPL}, they simultaneously learn domain-relevant knowledge in the domain-specific network. For instance, DMSN~\cite{yao2021multi} and TPKP~\cite{wu2022target} employ separate subnets for each source domain, while PMT~\cite{belal2024multi} and ACIA~\cite{belal2024attention} develop a prototype network and attention-based instance alignment method, respectively.
    
    \begin{figure}[t!]
        \centering
        \includegraphics[width=\linewidth]{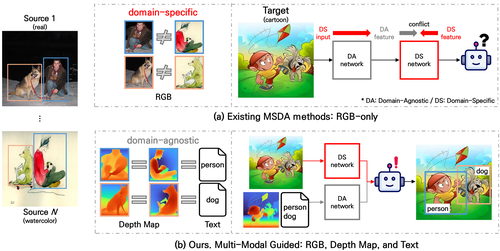}
        \caption{\textbf{Comparison of MSDA methods.}
        (a) Existing MSDA methods rely solely on RGB images to learn domain-agnostic features, while also encoding domain-specific features from them, which leads to a training conflict. (b) Our MS-DePro leverages additional domain-agnostic modalities—depth map and text—to encode domain-agnostic features.
        }
        \label{fig: intro}
    \end{figure}

    Although existing MSDA methods for OD have shown promising results, they have a fundamental limitation in modeling domain-agnostic and domain-specific knowledge. As shown in~\Cref{fig: intro}-(a), these methods first learn domain-invariant representations from domain-specific RGB images by intentionally suppressing domain-specific knowledge using GRL. From these domain-agnostic features, they then strive to encode domain-specific representations. This paradoxical process causes a training conflict between the objectives of these two networks, which impedes optimization and slows convergence. Furthermore, encoding domain-agnostic features from RGB images is fundamentally constrained, as they are highly sensitive to variations in illumination, color, and texture. To address this, we explicitly disentangle domain-specific and domain-agnostic features and extract them separately by leveraging additional domain-agnostic modalities. To this end, we consider depth maps and text (see~\Cref{fig: intro}-(b)). Compared to RGB images, depth maps provide more consistent representations by focusing on object shape and structure, thereby reducing reliance on domain-specific cues~\cite{gungor2024boosting}. Additionally, the text modality conveys high-level semantic information that remains consistent across domains~\cite{addepalli2024leveraging}. We argue that employing domain-agnostic modalities at the input level prevents interference from domain-specific knowledge.

    In this paper, we propose a novel framework termed Multi-Source Detector with Depth and Prompt (MS-DePro). Our MS-DePro consists of two main components: (1) depth-guided localization and (2) multi-modal guided text prompt learning. First, we exploit depth maps for domain-invariant object localization. To this end, we employ off-the-shelf monocular depth estimation models~\cite{depth_anything_v2, bochkovskii2024depth} to extract relative depth maps from input RGB images, which are only used during the training phase. The depth maps support RGB-only localization by complementing regions characterized by shape and structural cues. Next, multi-modal guided prompt learning leverages the high-level domain-agnostic characteristics of the text modality. Specifically, the domain-agnostic parts of the text prompt are enriched with features from depth maps, thus enhancing shared representations across multiple domains. Meanwhile, the domain-specific parts of the text prompt adjust text embeddings using RGB features, enabling domain-aware alignment between text and visual regions for improved object classification. Leveraging our proposed components, MS-DePro effectively encodes domain-agnostic features without relying on domain adversarial learning.

    The main contributions of this work are threefold:
    \begin{enumerate}
        \item We introduce MS-DePro, a novel MSDA framework for OD, which leverages auxiliary modalities, $i.e.,$ depth map and text, to explicitly encode domain-invariant representations.
        \item We propose depth-guided localization and multi-modal guided prompt learning. The depth-guided localization exploits depth maps to assist object localization, and multi-modal guided prompt learning explicitly integrates domain-agnostic and domain-specific features for improved classification.
        \item We conduct extensive experiments on MSDA benchmarks, and our MS-DePro achieves state-of-the-art performance. We further evaluate its generalization by extending experiments to the domain generalization setting.
    \end{enumerate}

    The remainder of this manuscript is organized as follows: \Cref{sec:related} provides a brief overview of the related work. \Cref{sec:method} elaborates the proposed MS-DePro and its two constituent modules. \Cref{sec:experiments} presents extensive experimental results on the MSDA and MSDG benchmarks and thorough ablation studies. \Cref{sec:conclusion} summarizes the findings and outlines the limitations.

\section{Related Work}
\label{sec:related}
    \subsection{Domain-Agnostic Representation Learning}
        As a branch of domain adaptation, research in multi-target~\cite{lu2025multiple}, blended-target~\cite{lu2026energy}, and source-free~\cite{lu2026exploring,lu2025adaptive} scenarios has focused on distribution alignment to minimize domain discrepancy. These diverse branches demonstrate that effectively mitigating domain shifts fundamentally relies on capturing domain-agnostic representations.
        Meanwhile, some studies~\cite{ouyang2022causality, song2025causality, zhou2025learning} have focused on learning causal representations to disentangle invariant features from spurious correlations through causal interventions within the RGB space, aiming to yield domain-agnostic representations. Specifically, Ouyang~et al.~\cite{ouyang2022causality} eliminated domain-specific appearance shortcuts, while Song~et al.~\cite{song2025causality} and Zhou~et al.~\cite{zhou2025learning} enforced causal separation through target style imitation and graph autoencoders, respectively. In this paper, we ensure the extraction of domain-agnostic features by leveraging depth maps and text directly at the input level, as RGB images are fundamentally domain-specific. By incorporating these inherently invariant modalities, our MS-DePro bypasses the limitations of relying solely on RGB inputs and enables the explicit capture of shared representations.
        
    \subsection{Multi-Source Domain Adaptation for Object Detection.}
        Multi-source domain adaptation (MSDA) for object detection assumes that training samples are collected from multiple sources. Following this assumption, several studies have aimed to adapt the model to a given target domain~\cite{yao2021multi, wu2022target, belal2024multi, belal2024attention}. For example, DMSN~\cite{yao2021multi} introduced a pseudo subnet to estimate target parameters by aggregating multi-source weights. TPKP~\cite{wu2022target} disentangled detection heads for each source domain to retain target-relevant knowledge. PMT~\cite{belal2024multi} performed class-conditioned alignment using domain-specific prototypes. ACIA~\cite{belal2024attention} proposed an additional attention-based class alignment coupled with an adversarial domain discriminator. All these methods were trained only on RGB images with strong domain-specific semantic cues, and attempted to learn domain-agnostic features implicitly through adversarial training or alignment-based techniques. In our method, we leverage additional input modalities, $i.e.,$ depth maps and text, both of which inherently contain domain-agnostic cues, to achieve domain-invariant localization and domain-tailored classification.

    \subsection{Depth-Assisted Object Detection.}
        Monocular depth estimation~\cite{depth_anything_v2, bochkovskii2024depth, wang2023sabv} estimates a relative or metric scale depth map from an RGB image. For example, Depth Anything~\cite{depth_anything_v2} and Depth Pro~\cite{bochkovskii2024depth}, trained on large-scale datasets, demonstrated notable zero-shot generalization performance. Recent studies leveraged depth maps for object detection, effectively reducing reliance on texture cues and preserving object shapes. For example, OVM3D-Det~\cite{huang2024training} utilized a depth estimation model to generate pseudo-LiDAR for open-vocabulary 3D object detection. WSOD-AMPLIFIER~\cite{gungor2024boosting} employed a Siamese network to learn hierarchical features from RGB and hallucinated depth maps for weakly supervised object detection. In this paper, we propose a depth-guided localization method that exploits depth maps and design a decoupled network for each domain to support RGB feature localization.

    \subsection{Prompt Learning.}
        Prompt learning has been widely studied to effectively leverage the knowledge encoded in pretrained VLMs for specific downstream tasks. CoOp~\cite{CoOp} introduced automated prompt engineering by replacing fixed context words with learnable continuous vectors. CoCoOp~\cite{CoCoOp} constructs a meta-token using a meta-network and appends it to the context tokens, thereby incorporating image-level features. DetPro~\cite{DetPro} extended CoOp for object detection by learning continuous prompt representations that distinguish the interpretation of the foreground and background. However, since those methods~\cite{CoOp, CoCoOp, DetPro} struggle with multiple domain scenarios, DAPL~\cite{DAPL} and DA-Pro~\cite{DA-Pro} separately learned domain-agnostic and domain-specific prompts for domain adaptation. While these works~\cite{DAPL, DA-Pro} have explored the separation of domain-agnostic and domain-specific components, they do not offer explicit guidance tailored to each part. In this paper, we propose multi-modal guided prompt learning to explicitly design domain-agnostic parts using depth map features and domain-specific parts using RGB features.

\section{Method}
\label{sec:method}
    \begin{figure*}[t!]
        \centering
        \includegraphics[width=\textwidth]{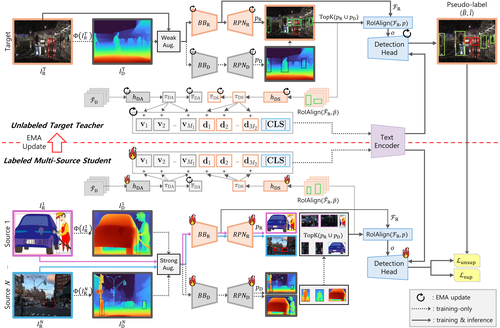}
        \caption{\textbf{Overall Framework of MS-DePro.}
            Following the mean-teacher framework, MS-DePro consists of an unlabeled target teacher (top) and a labeled multi-source student (bottom). We exploit depth maps to propose regions for localization, and our learnable prompt is encoded for classification. The teacher generates pseudo-labels for the target domain. The student learns from multiple sources and adapts to the target domain. The depth map is used only during training, while at inference time, only the teacher model with saved token buffers $\overline{\pi}_{\mathrm{DS}}$ and $\overline{\pi}_{\mathrm{DA}}$ is utilized.
            $\hat{\mathcal{F}}_\mathrm{R}$ and $\hat{\mathcal{F}}_\mathrm{D}$ denote the shallow feature maps from $BB_{\mathrm{R}}$ and $BB_{\mathrm{D}}$.
            }
        \label{fig: architecture}
    \end{figure*}
    
    \subsection{Problem Formulation}
        We introduce MS-DePro, a novel MSDA framework for OD. \Cref{fig: architecture} shows a summary of MS-DePro, which consists of two components: (1) depth-guided localization and (2) multi-modal guided prompt learning. In this section, we formalize the MSDA setting. Following that, we provide an overview of the model architecture. Then, in \Cref{subsec:depth_localization} and \Cref{subsec:multi_modal_guided_prompt_learning}, we describe the details of the proposed components.

        \noindent\textbf{MSDA Setting.}
             MSDA assumes access to $S$ source domains $\mathcal{D}^1, \mathcal{D}^2, \cdots, \mathcal{D}^S$ and a target domain $T$. Each $i$-th source domain $\mathcal{D}^{i}$ consists of $N^{j}$ labeled samples, where each sample includes an image $I^j$ and a set of annotations $\mathcal{A}^j$. Each $j$-th annotation is denoted as $(B, l)$, where $B$ denotes the bounding box, and $l$ is the corresponding label. For the target domain $T$, only $N^T$ images are available without any annotation.
    
        \noindent\textbf{Model Architecture.}
            The overall architecture of MS-DePro is depicted in \Cref{fig: architecture}. We use RegionCLIP~\cite{zhong2022regionclip}, a vision-language model based on Faster R-CNN~\cite{ren2016faster}. We initialized the backbone with pretrained ResNet-50~\cite{he2016deep} weights of RegionCLIP and the text encoder $\mathcal{T}(\cdot)$ with pretrained CLIP~\cite{radford2021learning} weights. Following the mean-teacher framework~\cite{tarvainen2017mean, li2022cross}, we first train the labeled multi-source student on $\mathcal{D}^1, \mathcal{D}^2, \cdots, \mathcal{D}^S$ source domains using a supervised loss $\mathcal{L}_\mathrm{sup}$, named the burn-in stage. In the domain adaptation stage, the unlabeled target teacher network is initialized with the exponential moving average (EMA) of the student weights. The teacher model generates pseudo-labels $(\overline{B}, \overline{l})$, which are then jointly optimized using an unsupervised loss ($\mathcal{L}_\mathrm{unsup}$) to train the student model. Note that the depth map is used only during training, while only the teacher model is used for inference.

    \subsection{Depth-Guided Localization}
    \label{subsec:depth_localization}
\begin{figure}[!t]
            \centering
            \includegraphics[width=\linewidth]{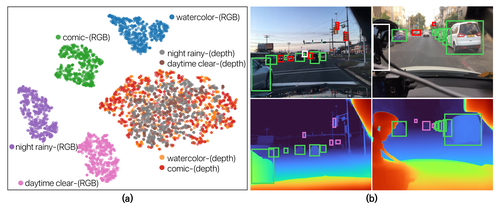}
            \caption{
                (a) t-SNE visualization of feature distribution for RGB and Depth map. (b) Region proposals on RGB and depth map images: green (common), white (spurious), red (RGB-only), and purple (depth-only).
            }
            \label{fig:depth_map_analysis}
        \end{figure}

        Our goal is to leverage depth maps and seamlessly integrate them into object localization to explicitly encode domain-agnostic features. This approach is motivated by the observation that models trained on RGB images tend to be biased toward domain-specific features, leading to shortcut learning~\cite{geirhos2020shortcut}. Although domain-specific features provide useful visual patterns for localization, this bias prevents the model from effectively capturing domain-agnostic cues, such as shape and structure, which are essential for precise localization. For instance, style cues\textemdash color and texture\textemdash are effective for localizing visually distinctive objects, yet they are highly sensitive to illumination changes ($e.g.,$ daytime vs. night). Therefore, relying solely on domain-specific features often degrades localization performance. To mitigate this, we exploit depth maps as an auxiliary domain-agnostic modality. As shown in \Cref{fig:depth_map_analysis}-(a), depth map features provide more consistent representations than RGB features. Further details are provided in~\Cref{subsec:depth_consistent}. Building on this, we introduce depth-guided localization, which complements domain-agnostic cues to guide the localization of RGB features.

        Concretely, given an input RGB image $I_{\mathrm{R}}^{i}$ from the $i$-th source domain, we estimate a relative depth map $I_{\mathrm{D}}^{i}$ using an off-the-shelf monocular depth estimation model~\cite{depth_anything_v2, bochkovskii2024depth}, formulated as $I_{\mathrm{D}}^{i} = \Phi(I_{\mathrm{R}}^{i})$. Here, $\Phi(\cdot)$ denotes the monocular depth estimation model. Then, $I_{\mathrm{R}}^{i}$ and $I_{\mathrm{D}}^{i}$ are fed into their own decoupled backbone ($BB_{\mathrm{R}}, BB_{\mathrm{D}}$) and region proposal network (RPN) ($RPN_{\mathrm{R}}, RPN_{\mathrm{D}}$). This design helps the model explicitly encode domain-specific and domain-agnostic features from RGB and depth maps, respectively. We then aggregate region proposals ${p_\mathrm{R}}$ and ${p_\mathrm{D}}$, generated from RGB and depth map RPN, respectively. In this step, we project the positional offsets of ${p_\mathrm{D}}$ onto the RGB feature map $\mathcal{F}_\mathrm{R}$ so that the model can encode regions that might be missed when using only RGB features. Subsequently, we select top-$k$ proposals with the highest objectness logits, defined as $p = \operatorname{TopK}({p_\mathrm{R} \cup p_\mathrm{D}})$. These selected proposals are encoded as object embeddings within the RGB feature map using RoIAlign~\cite{he2017mask}, formulated as $o = \mathrm{RoIAlign}(\mathcal{F}_\mathrm{R}, p)$.

        This framework is also employed by the teacher model for the unlabeled target domain to generate pseudo-labels. During the initial adaptation, we find that the model relying solely on domain-specific features struggles to generate reliable pseudo-labels due to the domain shift between the source and the unseen target domains. To compensate for this, we extract domain-agnostic regions from the depth map to refine ${p_\mathrm{R}}$ and generate complementary pseudo-boxes to improve adaptation. As depicted in \Cref{fig:depth_map_analysis}-(b), depth map proposals encode regions missed by RGB (purple) and filter out spurious ones (white). For common regions (green), we apply a weighted sum to yield a more precise offset.

        As shown in \Cref{fig: architecture}, the depth-guided localization network ($BB_{\mathrm{D}},~RPN_{\mathrm{D}}$) is structurally disentangled from the RGB network ($BB_{\mathrm{R}}, RPN_{\mathrm{R}}$) and operates exclusively during training, as indicated by dashed arrows. In this design, it provides auxiliary spatial guidance that encourages the RGB branch to learn domain-agnostic localization cues. At inference, the depth branch is simply removed without architectural modification, and detection is performed using the RGB network alone. This explicit separation enables a seamless transition between training and testing, while ensuring no additional memory or latency overhead during inference.

    \subsection{Multi-Modal Guided Prompt Learning}
    \label{subsec:multi_modal_guided_prompt_learning}
        \subsubsection{Preliminaries}  
            CLIP~\cite{radford2021learning} leverages a hand-crafted prompt, such as ``a photo of a [CLS]", to encode image-level text embeddings. However, since manual prompt tuning is time-consuming and often suboptimal, CoOp~\cite{CoOp} introduces a learnable context prompt for adaptive prompting. Specifically, the learnable prompt $\mathrm{t}_i$ for the $i$-th class consists of $M$ learnable tokens and is defined as follows:
            \begin{equation} \label{eq:CoOp}
            \begin{split}
               \mathrm{t}_i = [\mathrm{v}_1][\mathrm{v}_2]\dots[\mathrm{v}_M][\mathrm{CLS}_i],
            \end{split}
            \end{equation}
            where each $[\mathrm{v}_m] (m\in\{1,2,\dots, M\})$ represents a vector of the same dimension as text embeddings, and $\mathrm{CLS}_i$ denotes the tokenized vector.

        \subsubsection{Multi-Modal Guided Learnable Prompt} 
        \label{subsubsec: prompt}
            We leverage the text modality, which inherently provides domain-invariant representations through consistent high-level semantic information, to achieve robust object classification. We aim to incorporate domain-agnostic representations into the text prompt, capturing shared knowledge across multiple domains. Moreover, we consider domain-specific features to handle variations tailored to each domain’s unique characteristics.

            To this end, we propose a multi-modal guided learnable prompt that explicitly disentangles domain-specific and domain-agnostic knowledge, and captures both representations in a structured manner, as shown in the middle of \Cref{fig: architecture}. We decompose the $M$ learnable tokens in \Cref{eq:CoOp} into two sets: $[\mathrm{v}_m] \ (m \in {1, \dots, M_1})$ and $[\mathrm{d}_m] \ (m \in {1, \dots, M_2})$, designed to be combined with domain-agnostic and domain-specific information, respectively:
            \begin{equation} \label{eq:prompt_base}
            \begin{split}
               \mathrm{t}_i = [\mathrm{v}_{1}][\mathrm{v}_{2}]\dots[\mathrm{v}_{M_1}]
               [\mathrm{d}_{1}][\mathrm{d}_{2}]\dots[\mathrm{d}_{M_2}][\mathrm{CLS}_i].
            \end{split}
            \end{equation}
            First, we inject domain-specific characteristics extracted from the RGB images into the prompt, aiming to handle domain shifts. Following the previous studies~\cite{LowStyle1}, we consider shallow RGB features to contain domain-specific style cues. Unlike image classification, which utilizes holistic image-level information, object detection demands region-level understanding of both foreground and background. To address this, we extract region-level features by pooling shallow RGB features $\hat{\mathcal{F}}_\mathrm{R}$ using predicted region offsets. We then apply $\mathrm{RoIAlign}(\hat{\mathcal{F}}_\mathrm{R}, \hat{p})$, where $\hat{p}$ denotes the subset of top-$k$ proposals $p$ including all foreground and only 10\% of background proposals. In the target domain without ground-truth, objectness logits of proposals are employed. Subsequently, our domain-specific network (DS-Net) $h_\mathrm{DS}(\cdot)$ extracts a domain-specific token (DS token) from $\hat{\mathcal{F}}_\mathrm{R}$, computed as $\pi_{\mathrm{DS}} = h_{\mathrm{DS}}(\mathrm{RoIAlign}(\hat{\mathcal{F}}_\mathrm{R}, \hat{p}))$. The domain-specific parts of the prompt are then configured as $\mathrm{d}_{m}(I_\mathrm{R}) = \mathrm{d}_{m}+\pi_{\mathrm{DS}}$, where $m\in\{1,2,\dots,M_2\}$.

            Meanwhile, we focus on the depth modality, which provides domain-invariant geometric cues to facilitate the capture of domain-agnostic information shared across all domains. The domain-agnostic prompt is generated by injecting these domain-invariant attributes derived from shallow depth features $\hat{\mathcal{F}}_\mathrm{D}$ into learnable prompts. Specifically, given $\hat{\mathcal{F}}_\mathrm{D}$, our domain-agnostic network (DA-Net) $h_{\mathrm{DA}}(\cdot)$ extracts the domain-agnostic token (DA token) $\pi_{\mathrm{DA}}=h_{\mathrm{DA}}(\hat{\mathcal{F}}_\mathrm{D})$ and combines it with the $\mathrm{v}_{m}$, where $m\in\{1,2,\dots,M_1\}$. Consequently, our domain-agnostic parts of the prompt are represented as $\mathrm{v}_{m}(I_\mathrm{D}) = \mathrm{v}_{m}+\pi_{\mathrm{DA}}$.

            Taken together, our multi-modal guided learnable prompt is formulated as follows:
            \begin{equation} \label{eq:prompt_ours}
            \begin{split}
               \mathrm{t}_i(I_\mathrm{D}, I_\mathrm{R}) = &[\mathrm{v}_{1}(I_\mathrm{D})][\mathrm{v}_{2}(I_\mathrm{D})]\dots[\mathrm{v}_{M_1}(I_\mathrm{D})]\\
                              &[\mathrm{d}_{1}(I_\mathrm{R})][\mathrm{d}_{2}(I_\mathrm{R})]\dots[\mathrm{d}_{M_2}(I_\mathrm{R})][\mathrm{CLS}_i].
            \end{split}
            \end{equation}
            As the text representation ``background'' fails to capture the characteristics of background regions adequately, we allow the background category to learn its own embedding, additionally incorporating $\pi_{\mathrm{DA}}$ and $\pi_{\mathrm{DS}}$. Notably, the decomposed learnable token sets $[\mathrm{v}_m]$ and $[\mathrm{d}_m]$ are shared across all domains, ensuring that the total number of parameters remains constant as the number of domains increases.
    
    \subsection{Training and Inference}
        \subsubsection{Training Overview}
        \label{sec:training_overview}
            During training, we integrate our proposed prompt into the detection head. The prediction probability for class $y$ is calculated as the distance between the object embedding $o$ and the text embeddings $w_{i}=\mathcal{T}(t_i(I_\mathrm{D}, I_\mathrm{R})) \in \mathbb{R}^{Y \times \mathrm{d}}$:
            \begin{equation} \label{eq:cos}
            \begin{split}
                \hat{q}_y = \frac{\exp(s(o,w_y)/\tau)}{\sum_{i=1}^Y \exp(s(o,w_i)/\tau)},
            \end{split}
            \end{equation}
            where $s(\cdot,\cdot)$ is the cosine similarity, $Y$ denotes the number of classes along with a background, and $\tau$ is a temperature parameter. Finally, the predicted label is computed as $\hat{l} = \arg\max_{y \in \{1, 2, \dots, Y\}} \hat{q}_y$. The total training loss for the student model is:
            \begin{equation} \label{eq:total_loss}
            \begin{split}
                \mathcal{L} &= \mathcal{L}_{\text{sup}} + \alpha \mathcal{L}_{\text{unsup}} \\
                &= \left(\sum_{i=1}^{\mathcal{D}^{S}} \sum_{j=1}^{N^{i}} \mathcal{L}_\mathrm{cls}(\hat{l}_j^i, l_j^i) + \mathcal{L}_\mathrm{bbox}(\hat{b}_j^i, B_j^i) \right) \\
                & \quad + \alpha \left( \sum_{k=1}^{N^{T}} \mathcal{L}_\mathrm{cls}(\hat{l}_k^T, \overline{l}_k) + \mathcal{L}_\mathrm{bbox}(\hat{b}_k^T, \overline{B}_k) \right).
            \end{split}
            \end{equation}
            We employ focal loss~\cite{ross2017focal} for classification ($\mathcal{L}_{cls}$) and smooth-L1 loss~\cite{ren2016faster} for bounding box regression ($\mathcal{L}_{bbox}$), with $\hat{l}$ and $\hat{b}$ denoting the predicted label and bounding box. $\alpha$ is the hyperparameter for weighting the loss. The teacher's weights are updated via an EMA of the student's weights at each iteration, defined as $\theta_{t}^{(i)} = \lambda_{\alpha} \cdot \theta_{t}^{(i-1)} + (1 - \lambda_{\alpha}) \cdot \theta_{s}^{(i)}$, where $\theta_{t}$ and $\theta_{s}$ respectively denote the teacher and student model weights, $i$ is the current iteration, and $\lambda_{\alpha}$ is the weight smoothing coefficient. Similarly, the teacher's DA-Net, DS-Net, and the learnable token sets $[\mathrm{v}_m]$ and $[\mathrm{d}_m]$ are updated via EMA from their student counterparts with the same weight smoothing coefficient $\lambda_{\alpha}$.

        \subsubsection{Inference with Token Buffer} 
        \label{subsubsec: meta-token buffer}
            As mentioned in \Cref{subsec:depth_localization}, since depth maps are only used during training, the DA token $\pi_{\mathrm{DA}}$ cannot be constructed at inference. Therefore, we pre-construct a DA token buffer $\overline{\pi}_{\mathrm{DA}}$ during training to replace $\pi_{\mathrm{DA}}$ at inference. To this end, $\overline{\pi}_{\mathrm{DA}}$ is updated via EMA, following $\overline{\pi}_{\mathrm{DA}}^{\text{iter}} = \lambda_{\beta}\overline{\pi}_{\mathrm{DA}}^{\text{iter}-1} + (1-\lambda_{\beta})\pi_{\mathrm{DA}}^{\text{iter}}$, at each training iteration. In the target adaptation phase, we configure the DS token buffer $\overline{\pi}_{\mathrm{DS}}$ in a similar manner to the approach described above. Note that while the DA token $\pi_{\mathrm{DA}}$ is obtained from all domains, the DS token $\pi_{\mathrm{DS}}$ is derived from the target domain. Leveraging these buffers, we can pre-compute an offline text prompt for inference, as follows:
            \begin{equation} \label{eq:prompt_buffer}
            \begin{split}
               \overline{\mathrm{t}}_i = 
               [\overline{\mathrm{v}}_{1}][\overline{\mathrm{v}}_{2}]\dots[\overline{\mathrm{v}}_{M_1}][\overline{\mathrm{d}}_{1}][\overline{\mathrm{d}}_{2}]\dots[\overline{\mathrm{d}}_{M_2}][\mathrm{CLS}_i],
            \end{split}
            \end{equation}
            where $\overline{\mathrm{d}}_{m} = \mathrm{d}_{m}+\overline{\pi}_{\mathrm{DS}}$. Using the offline text prompt avoids calculating both the DA and DS tokens during inference, achieving an ensemble effect without introducing any additional computational cost. The probability is then computed by applying \Cref{eq:cos}, where the text embeddings $w_i$ are given by $\mathcal{T}(\overline{\mathrm{t}}_i)$.

\section{Experiments}
\label{sec:experiments}
    In this section, we evaluate MS-DePro on standard MSDA benchmarks. We also extend our experiments to a multi-source domain generalization benchmark to demonstrate the effect of explicitly modeling domain-invariant representations. Subsequently, we conduct ablations to analyze the contributions of the proposed components: (1) depth-guided localization and (2) multi-modal guided prompt learning.

    \noindent\textbf{Evaluation Metrics.}
        For all experiments, we evaluate performance with mAP@0.5 (mean Average Precision), computed at an IoU (Intersection over Union) threshold of 0.5.

    \noindent\textbf{Implementation Details.}
        We follow the experimental settings of~\cite{wu2022target, belal2024multi} for MSDA. For input images, we resized the shorter side to 600 pixels while maintaining aspect ratio, and all augmentations followed the methodology of~\cite{liu2021unbiased}. Here, RGB and depth map inputs were processed identically. We first trained our labeled multi-source student for 80k iterations in a supervised setting. Subsequently, we optimized for 80k iterations under both supervised and unsupervised settings. We used SGD for optimization, with a batch size of 8 for the source domain and 4 for the target domain, and a learning rate of 0.002. We set the EMA weight smoothing coefficients, $\lambda_{\alpha}$ and $\lambda_{\beta}$, to 0.9996 and 0.99, respectively. We configured the following hyperparameters: $\alpha$ was set to 1 for weighting loss; $M_1$ and $M_2$ were both set to 8 for the number of learnable tokens; and the pseudo-label confidence threshold was fixed at 0.9. All experiments were conducted on 4 NVIDIA A6000 GPUs, implemented using Detectron2~\cite{wu2019detectron2}.

    \subsection{Multi-Source Domain Adaptation}
    \label{sec:msda_benchmark}
    
        \noindent\textbf{Datasets.}
            For our MSDA experiments, we use 5 datasets: BDD100K~\cite{yu2020bdd100k}, Cityscapes~\cite{cordts2016cityscapes}, KITTI~\cite{geiger2012we}, MS-COCO~\cite{lin2014microsoft}, and Synscapes~\cite{wrenninge2018synscapes}.
            For more details about the datasets, please refer~\Cref{subsec:appendix_dataset_msda}.

        \noindent\textbf{Baselines.}
            We consider the following four baselines: (1) Lower Bound: Zero-shot and Source-only. In Source-only, RegionCLIP~\cite{zhong2022regionclip} is trained only with the source domain. (2) UDA: all source domains are aggregated into a single set, (3) MSDA: state-of-the-art (SOTA) MSDA methods designed for object detection, and (4) Oracle: Target-only and All-combined. In Target-only, RegionCLIP is trained with the labeled target domain. All-combined is trained with both source and labeled target domain datasets.

        \begin{table}[t]
            \centering
            \caption{Comparison of MSDA performance. Each column represents source domains, evaluated on subsets of BDD100K for cross-time (\textit{Dawn/Dusk}) and cross-camera (\textit{Daytime}).} 
            \label{tab: MSDA_time+camera}
            \begin{adjustbox}{width=\linewidth}
                \begin{tabular}{c l l c c c c c c}
                    \toprule
                    \multirow{2}{*}[-.3em]{Setting} & \multirow{2}{*}[-.3em]{Method} & \multirow{2}{*}[-.3em]{Venue} & \multicolumn{3}{c}{Cross-time} & \multicolumn{3}{c}{Cross-camera} \\
                    \cmidrule(lr){4-6}\cmidrule(lr){7-9}
                    & & & D & N & D+N & C & K & C+K \\
                    \midrule
                    \multirow{2}{*}{Lower Bound$^{*}$}
                    & \multicolumn{1}{l}{Zero-shot} & \multirow{2}{*}{CVPR'22} & - & - & 15.3 & - & - & 43.5 \\
                    & \multicolumn{1}{l}{Source-only} & & 42.8 & 32.9 & 36.1 & 47.2 & 41.9 & 45.2 \\
                    \midrule
                    \multirow{2}{*}{UDA}
                    & \multicolumn{1}{l}{UMT~\cite{deng2021unbiased}} & CVPR'21 & 33.8 & 21.6 & 33.5 & 47.5 & 35.4 & 47.0 \\
                    & \multicolumn{1}{l}{A.T.~\cite{li2022cross}} & CVPR'22 & 34.9 & 27.8 & 34.6 & 49.8 & 40.1 & 48.4 \\
                    \midrule
                    \multirow{7}{*}{MSDA} 
                    & \multicolumn{1}{l}{MDAN~\cite{zhao2018adversarial}} & NeurIPS'19 & - & - & 27.6 & - & - & 43.2 \\
                    & \multicolumn{1}{l}{M$^3$SDA~\cite{peng2019moment}} & ICCV'19 & - & - & 26.5 & - & - & 44.1 \\
                    & \multicolumn{1}{l}{DMSN~\cite{yao2021multi}} & ICCV'21 & - & - & 35.0 & - & - & 49.2 \\
                    & \multicolumn{1}{l}{TPKP~\cite{wu2022target}} & CVPR'22 & - & - & 39.8 & - & - & 58.4 \\
                    & \multicolumn{1}{l}{PMT~\cite{belal2024multi}} & WACV'24 & - & - & 45.3 & - & - & 58.7 \\
                    & \multicolumn{1}{l}{ACIA~\cite{belal2024attention}} & WACV'25 & - & - & 47.9 & - & - & 59.1 \\
                    & \multicolumn{1}{l}{MS-DePro (Ours)} & \multicolumn{1}{c}{KBS'26} &  - &  - &  \textbf{53.7} &  - &  - &  \textbf{68.4} \\
                    \midrule
                    \multirow{2}{*}{Oracle$^{*}$}
                    & \multicolumn{1}{l}{Target-only} & \multirow{2}{*}{CVPR'22} & - & - & 43.0 & - & - & 75.4 \\
                    & \multicolumn{1}{l}{All-combined} & & - & - & 50.8 & - & - & 76.0 \\
                    \bottomrule
                \end{tabular}
            \end{adjustbox}
            \begin{tablenotes}
                \item [] \scriptsize{* All results for Lower Bound and Oracle are evaluated using RegionCLIP.}
            \end{tablenotes}
        \end{table}
        
        \subsubsection{Cross-time and Cross-camera Adaptation} 
        \label{subsec: cross-time}
            
            \noindent\textbf{Setup.}
                For cross-time adaptation, we use the \textit{Daytime} (D) and \textit{Night} (N) subsets of BDD100K as source domains, while the \textit{Dawn/Dusk} (DD) subset of BDD100K serves as the target domain. As images are acquired at different times, changes such as illumination and object density cause a domain shift. Similarly, for cross-camera adaptation, we use Cityscapes (C) and KITTI (K) as source domains, with \textit{Daytime} as the target domain. In this setting, domain shift stems from different geographical locations and camera parameters. We conduct experiments on common categories between the source and target domains: 10 classes for cross-time and only \textit{car} for cross-camera.
    
            \noindent\textbf{Results.}
                UDA methods exhibit a slight performance drop when trained on multi-source domains compared to their single-source counterparts. This result implies that simply blending source domains leads to interference between multiple sources (D+N and C+K). In comparison, SOTA MSDA methods~\cite{belal2024multi, belal2024attention} show improved performance by modeling domain-specific features such as prototype and instance-level alignment. However, they rely solely on adversarial learning to learn domain-agnostic features. By contrast, we exploit additional depth maps and text modalities at the input level to design domain-invariant representations explicitly. Consequently, our MS-DePro achieves SOTA performance, outperforming ACIA~\cite{belal2024attention} by 5.8 mAP in cross-time and 9.3 mAP in cross-camera settings. It is noteworthy that MS-DePro is the first to outperform the Oracle All-combined baseline by 2.9 mAP in the cross-time setting.
    
        \begin{table}[t]
            \centering
            \caption{Comparison of MSDA performance in a mixed-domain setting, evaluated on \textit{Daytime} of BDD100K.} 
            \label{tab: MSDA_mix}
            \begin{adjustbox}{width=0.9\linewidth}
                \begin{tabular}{c l l c c c }
                    \toprule
                    \multirow{2}{*}[-.3em]{Setting} & \multirow{2}{*}[-.3em]{Method} & \multirow{2}{*}[-.3em]{Venue} & \multicolumn{3}{c}{Mixed-domain}\\
                    \cmidrule(){4-6}
                    & & & C & C+M & C+M+S \\
                    \midrule
                    \multirow{2}{*}{Lower Bound$^{*}$}
                    & \multicolumn{1}{l}{Zero-shot} & \multirow{2}{*}{CVPR'22} & - & - & 18.2  \\
                    & \multicolumn{1}{l}{Source-only} & & 32.6 & 34.1 & 34.9  \\
                    \midrule
                    \multirow{2}{*}{UDA}
                    & \multicolumn{1}{l}{UMT~\cite{deng2021unbiased}} & CVPR'21 & - & 29.7 & 30.9 \\
                    & \multicolumn{1}{l}{A.T.~\cite{li2022cross}} & CVPR'22 & - & 22.9 & 29.6 \\
                    \midrule
                    \multirow{4}{*}{MSDA}
                    & \multicolumn{1}{l}{TPKP~\cite{wu2022target}} & CVPR'22 & - & 35.3 & 37.1 \\
                    & \multicolumn{1}{l}{PMT~\cite{belal2024multi}} & WACV'24 & - & 38.7 & 39.7 \\
                    & \multicolumn{1}{l}{ACIA~\cite{belal2024attention}} & WACV'25 & - & 41.0 & 42.3 \\
                    & \multicolumn{1}{l}{ MS-DePro (Ours)} & \multicolumn{1}{c}{KBS'26} &  - &  \textbf{43.9} &  \textbf{46.7}\\
                    \midrule
                    \multirow{2}{*}{Oracle$^{*}$} 
                    & \multicolumn{1}{l}{Target-only} & \multirow{2}{*}{CVPR'22} & - & - & 53.3 \\
                    & \multicolumn{1}{l}{All-combined} & & - & 53.8 & 54.2 \\
                    \bottomrule
                \end{tabular}
            \end{adjustbox}
            \begin{tablenotes}
                \item [] \scriptsize{* All results for Lower Bound and Oracle are evaluated using RegionCLIP.}
            \end{tablenotes}
        \end{table}

        \subsubsection{Mixed-domain Adaptation}
            \noindent\textbf{Setup.}
                We employ the Cityscapes (C), MS-COCO (M), and Synscapes (S) datasets as source domains, while the \textit{Daytime} subset of BDD100K serves as the target domain. In this setting, the number of source domains is increased from two (C+M) to three (C+M+S) to simulate more challenging domain discrepancies. A domain shift mainly results from variations in scene (indoor vs. outdoor) and style (real vs. synthetic). We train and evaluate on 7 categories common to the sources and a target domain.
    
            \noindent\textbf{Results.}
                \Cref{tab: MSDA_mix} presents the results. We observe a consistent performance gain across all methods, even with the increased complexity introduced by the synthetic dataset (S). MS-DePro achieves SOTA performance, surpassing the previous SOTA method~\cite{belal2024attention} by 2.9 mAP in the C+M and 4.4 mAP in the C+M+S setting. We argue that domain-agnostic features play a key role in mitigating domain shift, particularly when the number of source domains increases.

    \subsection{Extension to Domain Generalization}
        Although MS-DePro is fundamentally developed for MSDA, we also extend our experiments to the multi-source domain generalization (MSDG) for object detection to evaluate how well our designed domain-invariant representations generalize to unseen domains. In the MSDG setting, where a DS token buffer $\overline{\pi}_{\mathrm{DS}}$ for the target domain cannot be pre-constructed, we compute the text embedding $w_i=\mathcal{T}(\overline{\mathrm{t}}_i(I_\mathrm{R}))$ using the text prompt described below:
        \begin{equation} \label{eq:prompt_buffer_MSDG}
        \begin{split}
           \overline{\mathrm{t}}_i(I_\mathrm{R}) = 
           &[\overline{\mathrm{v}}_{1}][\overline{\mathrm{v}}_{2}]\dots[\overline{\mathrm{v}}_{M_1}] \\
           &[\mathrm{d}_{1}(I_\mathrm{R})][\mathrm{d}_{2}(I_\mathrm{R})]\dots[\mathrm{d}_{M_2}(I_\mathrm{R})][\mathrm{CLS}_i].
        \end{split}
        \end{equation}
        Note that only the burn-in stage is conducted without the adaptation stage, and inference is performed using only the student model.

        \begin{table}[t]
            \centering
            \caption{Comparison of MSDG performance on Diverse Weather Dataset. Highest performance in bold, second-highest underlined.}
            \label{tab: MSDG_DWD}
            \begin{adjustbox}{width=0.9\linewidth}
                \begin{tabular}{ l l c c c c c c }
                    \toprule
                    \multirow{2}{*}{Method} & \multirow{2}{*}{Venue} & \multicolumn{2}{c}{Source} & \multicolumn{3}{c}{Unseen Target} & \multirow{2}{*}{mPC*} \\
                    \cmidrule(lr){3-4}\cmidrule(lr){5-7}
                     &  & DS & NC & DF & DR & NR &  \\
                    \midrule
                    CLIP-Gap~\cite{vidit2023clip} & CVPR'23 & 45.0 & 43.9 & 28.8 & 21.9 & 16.0 & 22.2 \\
                    PDOC~\cite{li2024prompt} & CVPR'24 & 47.1 & 46.3 & 32.0 & 24.4 & 16.9 & 24.4 \\ 
                    DivAlign~\cite{danish2024improving} & CVPR'24 & 46.5 & 47.7 & 34.2 & 34.9 & \underline{29.2} & \underline{32.7} \\
                    G-NAS~\cite{wu2024g} & AAAI'24 & 51.4 & 48.0 & 32.4 & 30.1 & 22.8 & 28.4 \\
                    OA-DG~\cite{lee2024object} & AAAI'24 & 54.1 & 36.6 & 36.0 & 32.0 & 18.7 & 28.9 \\
                    PhysAug~\cite{xu2025physaug} & AAAI'25 & 51.0 & 45.7 & 34.9 & \underline{35.1} & 27.1 & 32.4 \\
                    Diff. Guided~\cite{he2025generalized} & CVPR'25 & \underline{54.3} & \underline{49.3} & \textbf{39.7} & 32.4 & 25.6 & 32.6 \\
                    \midrule
                     MS-DePro (Ours) & \multicolumn{1}{c}{KBS'26} &  \textbf{55.5} &  \textbf{62.8} &  \underline{38.3} &  \textbf{35.2} &  \textbf{29.7} &  \textbf{34.4} \\
                    \bottomrule
                \end{tabular}
            \end{adjustbox}
            \begin{tablenotes}
                \item [] \scriptsize{* mPC: Average mAP across unseen target domains.}
            \end{tablenotes}
        \end{table}

        \noindent\textbf{Datasets.}
            In our experiments, we utilize 9 datasets: BDD100k~\cite{yu2020bdd100k}, FoggyCityscapes \cite{sakaridis2018semantic}, AdverseWeather~\cite{hassaballah2020vehicle}, VDD-DAOD~\cite{wu2021vector}, MS COCO~\cite{lin2014microsoft}, Pascal VOC~\cite{Everingham15}, Clipart1k (CP), Comic2k (CM) and Watercolor2k (WC)~\cite{inoue2018cross}.
            Please refer to~\Cref{subsec:appendix_dataset_msdg} for more details about the dataset.

        \noindent\textbf{Baseline.}
            We re-implement SOTA single-source domain generalization for object detection (S-DGOD) methods in the MSDG setting, using ResNet-50~\cite{he2016deep} as a backbone for a fair comparison with our method.
                    
        \subsubsection{Diverse Weather Domain Generalization}
            \noindent\textbf{Setup.}
                The conventional S-DGOD follows~\cite{inoue2018cross, wu2022single}, training on DS and then evaluating on the unseen domains (NC, DF, DR, and NR). We extend this setting to MSDG setting, performing training on DS and NC (DS+NC) and then evaluating on DF, DR, and NR. We train and evaluate on 7 categories shared across all source and target domains.

            \noindent\textbf{Results.}
                In~\Cref{tab: MSDG_DWD}, MS-DePro achieves the best performance on all source (seen) and unseen domains—except DF, where it ranks second—as well as on the mPC metric, which averages mAP over the unseen domains. Among previous methods, only PDOC~\cite{li2024prompt} and CLIP-Gap~\cite{vidit2023clip} utilize text. However, their text prompts implicitly assume target domain knowledge is available for unseen domains ($e.g.,$ ``an image taken on a \textit{fog day}'', ``an image taken on a \textit{rain night}''). In contrast, our learnable prompts integrate domain knowledge from input modalities. Moreover, MS-DePro employs explicit feature separation to encode domain-relevant and domain-invariant representations, in comparison with prior methods, which mainly rely on image augmentation.

        \begin{table}[t]
            \centering
            \caption{Comparison of MSDG performance on Real to Artistic.}
            \label{tab:MSDG_cross_domain}
            \begin{adjustbox}{width=0.95\linewidth}
                \begin{tabular}{ l l c c c c c c}
                    \toprule
                    \multirow{2}{*}{Method} & \multirow{2}{*}{Venue} & \multicolumn{2}{c}{Source} & \multicolumn{3}{c}{Unseen Target} & \multirow{2}{*}{mPC**} \\
                    \cmidrule(lr){3-4}\cmidrule(lr){5-7}
                     &  & VOC & COCO* & Clipart & Comic & Watercolor &  \\
                    \midrule
                    CLIP-Gap~\cite{vidit2023clip} & CVPR'23 & - & - & - & 32.7 & 48.1 & - \\
                    PDOC~\cite{li2024prompt} & CVPR'24 & 76.4 & - & 16.7 & 27.2 & 41.7 & 28.5 \\
                    OA-DG~\cite{lee2024object} & AAAI'24 & 78.7 & - & 29.7 & 20.4 & 44.0 & 31.4 \\
                    \midrule
                     MS-DePro (Ours) & \multicolumn{1}{c}{KBS'26} & \textbf{78.8} & - & \textbf{47.5} &  \textbf{42.5} & \textbf{57.5} & \textbf{49.2} \\
                    \bottomrule
                \end{tabular}
            \end{adjustbox}
            \begin{tablenotes}
                \item [] \scriptsize{* MS COCO is not evaluated, being used only for training.}
                \item [] \scriptsize{** mPC: Average mAP across unseen target domains.}
            \end{tablenotes}
        \end{table}

        \subsubsection{Real to Artistic Domain Generalization}
            \noindent \textbf{Setup.}
                By default, real to artistic generalization is trained on a real-world domain~\cite{Everingham15} and evaluated on multiple artistic domains~\cite{inoue2018cross} to assess generalization on the real-world to artistic domain gap. To preserve the experimental assumption that real and artistic domains remain distinct, we additionally include MS COCO in the source domain (COCO), extending it into a multi-source setting.
                
            \noindent\textbf{Results.}
                \Cref{tab:MSDG_cross_domain} shows the results. We benchmark SOTA S-DGOD methods in the real to artistic setting and observe that our MS-DePro consistently outperforms them. In particular, CLIP-Gap~\cite{vidit2023clip} and PDOC~\cite{li2024prompt} exhibited limited performance, even though they used art-domain prompts ($e.g.$, ``an image in the \textit{comics} style'') that contradict the experimental assumption of having no prior knowledge of the unseen target domain. In contrast, our MS-DePro explicitly leverages auxiliary domain-agnostic modalities to learn domain-invariant features at the input level.

        \begin{table}[t]
            \centering
            \caption{Comparison of performance and computational cost.} 
            \label{tab:fair_comparison}
            \begin{adjustbox}{width=\linewidth}
                \begin{threeparttable}
                \begin{tabular}{l l l c c c c c c c c c}
                    \toprule
                    \multirow{3}{*}[-.3em]{Method} & \multirow{3}{*}[-.3em]{Backbone} & \multirow{3}{*}[-.3em]{Weight} & \multirow{3}{*}[-.3em]{Depth} & \multirow{3}{*}[-.3em]{Text\tnote{1}} & \multirow{3}{*}[-.3em]{\makecell{mAP \\ (\%)}} & \multicolumn{2}{c}{Train} & \multicolumn{4}{c}{Test} \\
                    \cmidrule(lr){7-8}\cmidrule(lr){9-12}
                    & & & & & & \multicolumn{2}{c}{\#Param (M)} & \multicolumn{2}{c}{\#Param (M)} & \multirow{2}{*}{\makecell{FLOPs \\ (G)}} & \multirow{2}{*}{\makecell{Latency \\ (ms)}} \\
                    & & & & & & BB.\tnote{2} & Det.\tnote{2} & BB.\tnote{2} & Det.\tnote{2} & & \\
                    \midrule
                    \multirow{3}{*}[-.3em]{\makecell[l]{Baseline \\ (Faster R-CNN)}} & VGG-16 & ImageNet & - & - & 44.9 & 14.7 & 29.2 & 14.7 & 29.2 & 223.6 & 36.1 \\
                    \cmidrule{2-12}
                    & \multirow{2}{*}{ResNet-50} & ImageNet & - & - & 45.4 & 23.3 & 24.5 & 23.3 & 24.5 & 267.0 & 42.2 \\
                    & & RegionCLIP & - & -  & 47.7 & 23.3 & 24.5 & 23.3 & 24.5 & 267.0 & 42.2 \\
                    \midrule
                    ACIA~\cite{belal2024attention} & VGG-16 & ImageNet & - & - & 47.9 & 14.7 & 29.2 & 14.7 & 29.2 & 223.6 & 36.1 \\
                    \midrule
                    \multirow{4}{*}[-.3em]{\makecell[l]{MS-DePro \\ (Ours)}} & VGG-16 & ImageNet & \checkmark & - & 48.2 & 29.4 & 31.6 & 14.7 & 29.2 & 223.6 & 36.1 \\
                    \cmidrule{2-12}
                    & \multirow{3}{*}[-.3em]{ResNet-50} & ImageNet & \checkmark & - & 48.8 & 46.6 & 34.0 & 23.3 & 24.5 & 267.0 & 42.2 \\
                    \cmidrule{3-12}
                    &  & \multirow{2}{*}{RegionCLIP} & \checkmark & - & 49.7 & 46.6 & 34.0 & 23.3 & 24.5 & 267.0 & 42.2 \\
                    & & &  \checkmark &  \checkmark &  53.7 &  46.6 &  34.1 &  23.3 &  24.5 &  267.0 &  42.4 \\
                    \bottomrule
                \end{tabular}
                \begin{tablenotes}
                    \item [\tnote{1}] \normalsize{A dash (–) in text column indicates a linear classifier without the text modality.}
                    \item [\tnote{2}] \normalsize{BB. and Det. denote the backbone and detection modules ($e.g.,$ RPN and RoI Head), respectively.}
                    \item [*] \normalsize{Here, Depth and Text refer to depth-guided localization and multi-modal guided prompt learning, respectively.}
                \end{tablenotes}
                \end{threeparttable}
            \end{adjustbox}
        \end{table}

    \subsection{Ablation Studies}
    \label{sec:ablations}
        We present comprehensive ablation studies to analyze our proposed components. To this end, we conduct experiments under the cross-time setting in \Cref{subsec: cross-time}. The results demonstrate that the proposed components are effective and contribute synergistically to the model performance.

        \subsubsection{Impact of Backbone and Pretrained Weights.}
            Unlike previous MSDA methods that utilize the VGG-16 backbone, we adopt ResNet-50 backbone from RegionCLIP to incorporate the text modality. To ensure a fair comparison and validate that the performance gain comes from our proposed components rather than the backbone change, we analyze the impact of different backbones and pretrained weights in~\Cref{tab:fair_comparison}. First, ResNet-50 weights of RegionCLIP~\cite{zhong2022regionclip} are pretrained on the large-scale CC3M dataset~\cite{sharma2018conceptual}, which is larger than ImageNet and shows higher baseline performance (Baseline in~\Cref{tab:fair_comparison}). Nevertheless, it is worth noting that the CC3M dataset does not overlap with any target domains used in our experiments.
            
            With the VGG-16 backbone, our method with depth-guided localization (48.2 mAP) surpasses both ACIA (47.9 mAP) and the baseline (44.9 mAP). Moreover, when using the ResNet-50 backbone with RegionCLIP weights, MS-DePro significantly outperforms the baseline (53.7 vs. 47.7 mAP). Notably, this is a substantial improvement over our model variant without text (49.7 mAP), which confirms the strong contribution of our multi-modal guided prompt learning.
            
        \subsubsection{Comparison of Computational Cost.}
            In~\Cref{tab:fair_comparison}, we also report the number of parameters (\#Param), FLOPs, and latency. Although the proposed depth-guided localization introduces extra training costs, it is applied exclusively during training and removed during inference. As the comparison between our baseline (row 3) and the model with depth-guided localization (row 7) shows, the number of Parameters, FLOPs, and Latency at test time are identical. Moreover, while our multi-modal guided prompt learning introduces only negligible computational overhead, it delivers a substantial performance gain (49.7 $\rightarrow$ 53.7 mAP). Finally, compared to ACIA, our MS-DePro incurs a slight increase in test time computational cost (223.6G $\rightarrow$ 267.0G FLOPs; 36.1ms $\rightarrow$ 42.4ms latency) but achieves a significant performance improvement of +5.8 mAP.

        \subsubsection{Effects of the Main Components.}
            \Cref{tab: ablation_main} shows the effects of our main contributions. Using a linear classifier (Exp.~\#1 and Exp.~\#2), depth guidance provides a 2.0\% gain in mAP, while employing a text-guided classifier (Exp.~\#3 \textemdash \#6) achieves consistently higher performance. For depth-guided localization, we observe performance gains of 1.1\% (Exp.~\#3 $\rightarrow$ \#4) and 1.6\% (Exp.~\#5 $\rightarrow$ \#6) for fixed and learnable prompt types, respectively. The usage of a learnable prompt achieves higher performance than a fixed prompt (Exp.~\#3 vs. Exp.~\#5). Note that when the depth map is not used, only $\pi_{\mathrm{DS}}$ is applied in Exp.~\#5. Consequently, the combination of these components yields synergistic gains, outperforming their individual effects (Exp.~\#6).

        \subsubsection{Ablation on Diverse Geometric Cues.}
        \label{sec:ablations_geo_refine}
            In \Cref{tab: ablation_depth}, we evaluate the effect of various geometric cues. To this end, we exploit three models to encode depth and normal maps, \textit{i.e.}, Depth Pro~\cite{bochkovskii2024depth}, Depth Anything V2~\cite{depth_anything_v2}, and Omnidata~\cite{eftekhar2021omnidata}. We confirm that leveraging geometric cues consistently improves performance compared to RGB-only (+0.8, +1.0, and +1.6 mAP). These results suggest that using geometric cues enhances the quality of region proposal and pseudo-label in the domain adaptation stage.

        \begin{table}[t]
            \centering
            \caption{Ablation on main components of MS-DePro. $\pi_{\mathrm{DA}}$ and $\pi_{\mathrm{DS}}$ respectively indicate DA and DS token.} 
            \label{tab: ablation_main}
            \begin{adjustbox}{width=0.75\linewidth}
                \begin{tabular}{c c c c c c}
                    \toprule
                    \multirow{2}{*}[-.3em]{Exp.~\#} & \multicolumn{1}{c}{\multirow{2}{*}[-.3em]{Depth-Guided Loc.}} & \multicolumn{1}{c}{\multirow{2}{*}[-.3em]{Classifier}} & \multicolumn{2}{c}{Prompt} & \multirow{2}{*}[-.3em]{mAP} \\
                    \cmidrule(lr){4-5}
                    &  &  & \multicolumn{1}{c}{Type} & \multicolumn{1}{c}{Token} &  \\
                    \midrule
                    1 & - & \multirow{2}{*}{linear} & - & - & 47.7 \\
                    2 & \checkmark & & - & - & 49.7 \\
                    \midrule
                    3 & - & \multirow{4}{*}{text} & fix & - & 51.3 \\
                    4 & \checkmark & & fix & - & 52.4 \\
                    5 & - & & learnable & $\pi_{\mathrm{DS}}$ & 52.1 \\
                    6 & \checkmark & & learnable & $\pi_{\mathrm{DA}}, \pi_{\mathrm{DS}}$ & \textbf{53.7} \\
                    \bottomrule
                \end{tabular}
            \end{adjustbox}  
        \end{table}
        
        \begin{table}[t]
        \centering
            \caption{Ablation on diverse geometric cues.}
            \label{tab: ablation_depth}
            \begin{adjustbox}{width=0.5\linewidth}
            \begin{tabular}{ l l c}
                \toprule
                Geometric & Method & mAP \\
                \midrule
                -      & RGB-only                                   & 52.1 \\
                \midrule
                Normal & Omnidata~\cite{eftekhar2021omnidata}       & 52.9 \\
                Depth  & Depth Anything V2~\cite{depth_anything_v2} & 53.1 \\
                Depth  & Depth Pro~\cite{bochkovskii2024depth}      & \textbf{53.7} \\
                \bottomrule
            \end{tabular}
            \end{adjustbox}
        \end{table}

        \begin{table}[t]
            \centering
            \caption{Ablation on the configuration of DA token and DS token.} 
            \label{tab: ablation_meta-Token}
            \begin{adjustbox}{width=0.6\linewidth}
                \begin{tabular}{c c c c}
                    \toprule
                    \multirow{2}{*}[-.3em]{Exp. \#} & \multicolumn{2}{c}{Additional-Token} & \multirow{2}{*}[-.3em]{mAP}\\
                    \cmidrule(lr){2-3}
                    & $\pi_{\mathrm{DA}}$ & \multicolumn{1}{c}{$\pi_{\mathrm{DS}}$} & \\
                    \midrule
                    1 &- & - & 50.6 \\
                    2 &- & Proposal-level& 52.4 \\
                    3 &Image-level & - & 52.7 \\
                    4 &Proposal-level & Proposal-level & 52.2 \\
                    5 &Image-level & Image-level & 52.8 \\
                    6 &Image-level & Proposal-level & \textbf{53.7} \\
                    \bottomrule
                \end{tabular}
            \end{adjustbox}
        \end{table}
            
        \begin{table}[t]
            \centering
            \caption{Ablation on different prompt templates.} 
            \label{tab: ablation_prompt}
            \begin{adjustbox}{width=0.7\linewidth}
                \begin{tabular}{l l c}
                    \toprule
                    Method & Template & mAP \\
                    \midrule
                    CLIP~\cite{radford2021learning} & ``a photo of a $[\mathrm{CLS}]$''& 52.4 \\
                    CoOp~\cite{CoOp} & ``$[\mathrm{v}_{1}][\mathrm{v}_{2}]\dots[\mathrm{v}_{M}] [\mathrm{CLS}]$'' & 50.6 \\
                    CoCoOp~\cite{CoCoOp} & ``$[\mathrm{v}_{1}(I_\mathrm{R})][\mathrm{v}_{2}(I_\mathrm{R})]\dots[\mathrm{v}_{M}(I_\mathrm{R})] [\mathrm{CLS}]$'' & 51.1 \\
                    DAPL~\cite{DAPL} & ``$[\mathrm{v}_{1}]\dots[\mathrm{v}_{M_1}] [\mathrm{d}_{1}^{d}]\dots[\mathrm{d}_{M_2}^{d}][\mathrm{CLS}]$'' & 47.8 \\
                     Ours &``$[\overline{\mathrm{v}}_{1}]\dots[\overline{\mathrm{v}}_{M_1}][\overline{\mathrm{d}}_{1}]\dots[\overline{\mathrm{d}}_{M_2}][\mathrm{CLS}]$''  & \textbf{53.7} \\
                    \bottomrule
                \end{tabular}
            \end{adjustbox}
        \end{table}

        \subsubsection{Where do Features for Tokens come from?}
            As described in \Cref{subsubsec: prompt}, we construct two types of additional tokens, $\pi_{\mathrm{DA}}$ and $\pi_{\mathrm{DS}}$. We conduct an empirical experiment to determine the optimal set of additional tokens in \Cref{tab: ablation_meta-Token}. Incorporating tokens generated from shallow image-level RGB and depth features (Exp.~\#5) improves performance to 52.8 mAP, showing a 2.2 mAP gain over Exp.~\#1, which does not apply any tokens (same as CoOp~\cite{CoOp} in \Cref{eq:CoOp}). However, as object detection inherently predicts the class of regions, we consider that domain-specific style information in these regions is crucial. Thus, we extract the DS token at the proposal-level, achieving our best performance, as presented in Exp.~\#6. A comparison between Exp.~\#4 and Exp.~\#6 shows that incorporating global geometric information at the image level, rather than the proposal level, facilitates more effective domain-agnostic representations.

        \subsubsection{Comparison with other Prompting Strategies.}
            We compare our multi-modal guided learnable prompt with other prompting studies, as illustrated in \Cref{tab: ablation_prompt}. To ensure a fair comparison, we use the same network while varying only the prompting strategy. While CoOp~\cite{CoOp} underperforms CLIP~\cite{radford2021learning} on the target domain, it yields a 0.7 higher mAP under daytime conditions that share a similar distribution with the source domain. This discrepancy suggests that simply learning prompts without any feature conditioning and explicit guidance leads to overfitting on the source domains. DAPL~\cite{DAPL} considers domain-specific attributes but still performs worse than CLIP, as it shares the same limitation as CoOp, namely learning prompts directly without feature guidance. Moreover, since its domain-specific tokens ($[\mathrm{d}_{1}^{d}] \dots [\mathrm{d}_{M_{2}}^{d}]$) are constructed independently for each domain, they become biased toward unreliable pseudo-labels during the early target adaptation stage. On the other hand, CoCoOp~\cite{CoCoOp} guides each token with RGB deep features but lacks a design suited for domain adaptation and object detection, resulting in 51.1 mAP. Unlike previous approaches, we leverage depth and RGB features to guide domain-agnostic and domain-specific tokens, respectively. We achieve 53.7 mAP, the highest among all methods, highlighting the effectiveness of our prompting strategy for MSDA.
    
        \subsubsection{Impact of Learnable Prompt Token Lengths}
            We investigate the impact of the number of learnable prompt tokens, $M_1$ and $M_2$, to identify the optimal balance between domain-agnostic and domain-specific components. As shown in~\Cref{tab: prompt_m1m2_parameters}, a balanced configuration of $M_1=8$ and $M_2=8$ achieves the highest performance with 53.7 mAP. Notably, allocating the entire token budget to either the domain-specific ($M_1=16$) or domain-agnostic ($M_2=16$) component alone results in a performance decline. This suggests that the synergistic interaction between both types of prompts is more effective than relying on a single attribute for representation. Furthermore, increasing the total tokens to 32 ($M_1=16, M_2=16$) provides no additional gains. Consistent with the findings in DA-Pro~\cite{DA-Pro}, this suggests that excessive prompt lengths cause overfitting to source-specific characteristics, thereby hindering effective adaptation to target domains.

        \begin{table}[t]
            \centering
            \caption{Ablation on the number of learnable prompt tokens.} 
            \label{tab: prompt_m1m2_parameters}
            \begin{adjustbox}{width=0.65\linewidth}
                \begin{tabular}{l c c c}
                    \toprule
                    Template &$M_1$ & $M_2$ &  mAP \\
                    \midrule
                    \multirow{4}{*}{``$[\overline{\mathrm{v}}_{1}]\dots[\overline{\mathrm{v}}_{M_1}][\overline{\mathrm{d}}_{1}]\dots[\overline{\mathrm{d}}_{M_2}][\mathrm{CLS}]$''} & 8 & 8 & \textbf{53.7} \\
                    & 0 & 16 & 52.7 \\
                    & 16 & 0 & 53.1 \\
                    & 16 & 16 & 53.3 \\
                    \bottomrule
                \end{tabular}
            \end{adjustbox}
        \end{table}
        
        \begin{table}[t]
            \centering
            \caption{Ablation on objectness logits (\%) for constructing target DS token $\pi_{\mathrm{DS}}$ and pseudo-labels.} 
            \label{tab:objectness_logits}
            \begin{adjustbox}{width=0.35\linewidth}
                \begin{tabular}{ c c }
                    \toprule
                    Objectness Logits & mAP \\
                    \midrule
                    10\% & \textbf{53.7}  \\
                    30\% & 53.4   \\
                    50\% & 53.4   \\
                    100\% & 53.3   \\
                    \bottomrule
                \end{tabular}
            \end{adjustbox}
        \end{table}

    \begin{figure}[!t]
        \centering
        \includegraphics[width=0.75\linewidth]{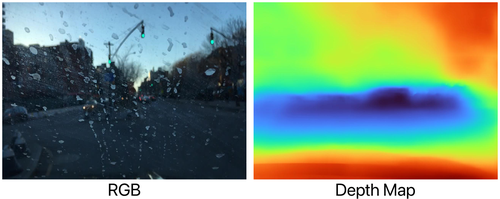}
        \caption{
            Visualization of an RGB image and its corresponding low-quality depth map, where raindrops on the windshield and low-light conditions resulted in degraded depth estimation.
        }
        \label{fig:low_quality_depth_map}
    \end{figure}

    \begin{figure}[!t]
        \centering
        \includegraphics[width=0.75\linewidth]{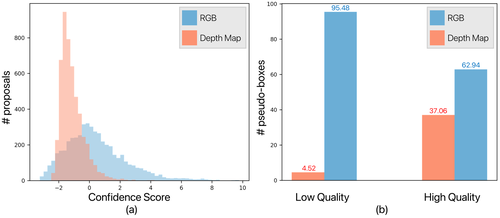}
        \caption{
            (a) Confidence scores of region proposals from RGB and depth maps. (b) Counts of pseudo-boxes from RGB and depth maps, with averages shown above the bars.
        }
        \label{fig:depth_map_vis_comparison}
    \end{figure}

    \subsubsection{Effective Pseudo-Labels for Target Adaptation.}
        In MS-DePro, the DS token $\pi_{\mathrm{DS}}$ is constructed from the entire foreground and 10\% of the background region to encode discriminative object-level characteristics across domains.
        In the case of the target domain, where ground-truth bounding boxes are unavailable, we cannot explicitly separate foreground and background regions.  Thus, we utilize only the top 10\% of proposals ranked by objectness logits during pseudo-label generation in MSDA for target adaptation.
        In~\Cref{tab:objectness_logits}, we analyze the impact of varying objectness logits on target adaptation performance. 
        As higher objectness logits are incorporated, more background regions contribute to the configuration of $\pi_{\mathrm{DS}}$ and pseudo-labels. 
        Consequently, selecting the top 10\% of objectness logits yield the highest performance at 53.7 mAP, whereas utilizing all proposals ($i.e.$, 100\% of objectness logits) leads to a slight performance degradation of 0.4 mAP.
        This indicates that appropriately selecting proposals based on objectness logits influences the quality of both the target DS token $\pi_{\mathrm{DS}}$ and the pseudo-labels.
        
    \subsection{Analysis}
    \label{sec:analysis}
        
        \subsubsection{Do depth maps have consistent representation?}
        \label{subsec:depth_consistent}
            As discussed in~\Cref{subsec:depth_localization}, we visualize feature distributions to investigate whether depth maps provide more consistent representations than RGB images (see~\Cref{fig:depth_map_analysis}-(a)). To this end, we employ four different domains, including adverse weather (Daytime Sunny, Night Rainy) and artistic style (Comic, Watercolor)~\cite{inoue2018cross}. These domains differ in lighting and weather conditions (Daytime Sunny vs. Night Rainy) and in style (real vs. artistic: Comic, Watercolor). As illustrated in the t-SNE visualization results of~\Cref{fig:depth_map_analysis}-(a), RGB features are clearly separated into domain-specific clusters, whereas depth features are aggregated into a single cluster. Moreover, we compute the average distribution distances of features. Specifically, we measure the distance between feature distributions using the Euclidean distance of their means. As a result, RGB features exhibit larger distances (40.558 for mean) compared to depth features (0.510 for mean), indicating that depth features remain closer across domains. These results suggest that depth maps are more consistent feature representations than RGB images.

        \subsubsection{In-Depth Analysis of Depth Map Quality}
        \label{subsec:analysis_depth}
            Although our MS-DePro achieves SOTA performance, the depth maps tend to be noisy under challenging conditions such as motion blur, severe occlusion, adverse weather, and poor lighting. Hence, a natural question arises: \textit{``How do low-quality depth maps affect model training?"}. To explore this, we conduct an in-depth analysis of low-quality depth maps and their impact on the model. As illustrated in~\Cref{fig:low_quality_depth_map}, we first sample 100 low-quality depth maps generated by a monocular depth estimation model~\cite{bochkovskii2024depth}. Specifically, we calculate entropy from depth histograms and cosine similarity between the gradient magnitudes of the RGB and the depth map. Then, depth maps whose quality score $Q$ is lower than a threshold are classified as low-quality samples. Next, we analyze the confidence scores of proposals generated from the sampled depth maps. As shown in~\Cref{fig:depth_map_vis_comparison}-(a), the proposals from low-quality depth maps show lower confidence scores than those from RGB images, and are mainly distributed between $-2$ and $0$. This indicates that the model mainly relies on RGB proposals when generating pseudo-labels, as also observed in~\Cref{fig:depth_map_vis_comparison}-(b). In~\Cref{fig:depth_map_vis_comparison}-(b), pseudo-labels generated from low-quality depth maps are markedly fewer than those from RGB (avg. 4.52 vs. 95.48), whereas high-quality depth maps provide considerably more pseudo-labels (avg. 37.08 vs. 62.94). From these results, we argue that low-quality depth maps only have a negligible impact on pseudo-label generation, while in most cases with high-quality depth maps, they provide meaningful guidance for model localization.

          \begin{table}[t]
                \centering
                \captionof{table}{Ablation study on DS token configuration. Class-wise frequency and performance drop relationship. Comparison of our class-agnostic (Exp.~\#1) and the class-specific strategies (Exp.~\#2).}
                \label{tab:DS_class-specific}
                \begin{adjustbox}{width=0.56\linewidth}
                \begin{tabular}{c c c c c}
                    \toprule
                    \multicolumn{2}{c}{Class} & \multicolumn{2}{c}{mAP} & \multirow{2}{*}[-.3em]{\makecell{Performance Gap \\ (Exp.~\#1 - Exp.~\#2)}} \\
                    \cmidrule(lr){1-2}\cmidrule(lr){3-4}
                    Name & Frequency & Exp.~\#1 & Exp.~\#2 & \\
                    \midrule
                    Car & 8649 & 77.3 & 75.9 & 1.4 \\
                    Sign & 2550 & 66.3 & 62.2 & 4.1 \\
                    Light & 2128 & 62.7 & 55.9 & 6.8 \\
                    Person & 1060 & 57.9 & 55.4 & 2.5 \\
                    Truck & 355 & 59.8 & 47.1 & 12.7 \\
                    Bus & 137 & 61.0 & 51.7 & 9.3 \\
                    Bike & 66 & 64.4 & 49.3 & 15.1 \\
                    Rider & 47 & 43.0 & 38.3 & 4.7 \\
                    Motor & 28 & 44.7 & 33.6 & 11.1 \\
                    Train & - & - & - & - \\
                    \midrule
                    \multicolumn{2}{c}{mAP} & \textbf{53.7} & 46.9 & 6.8 \\
                    \bottomrule
                \end{tabular}
            \end{adjustbox}  
            \vspace*{1.0\baselineskip}
        \end{table}
        
        \begin{figure}[!t]
            \centering
            \includegraphics[width=0.63\linewidth]{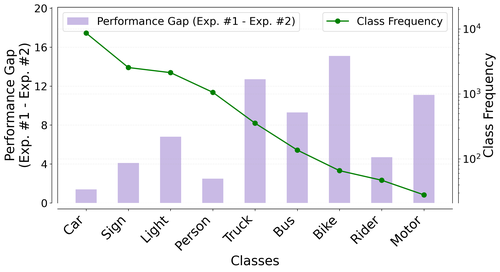}
            \caption{Impact of class frequency on performance gap. Under a class-specific configuration, rare classes ($e.g.,$ Bike, Motor) exhibit greater performance degradation, whereas frequent classes ($e.g.,$ Car, Sign) are less affected.}
            \label{fig:DS_class-specific}
        \end{figure}
  
        \subsubsection{Class-specific DS Token}
        \label{subsec:class_specific}        
            In~\Cref{subsec:multi_modal_guided_prompt_learning}, we configure a DS token $\pi_{\mathrm{DS}}$ based on the region offset of predicted foreground and background proposals. Intuitively, to effectively teach the model class-wise distinctions, $\pi_{\mathrm{DS}}$ can be constructed in a class-specific manner by assigning the class for each proposal. However, configuring $\pi_{\mathrm{DS}}$ in a class-specific manner results in a significant performance drop (see \Cref{tab:DS_class-specific} and \Cref{fig:DS_class-specific}). Exp.~\#1 corresponds to our original model (class-agnostic approach), while Exp.~\#2 adopts the class-specific approach. Notably, we observe an inverse correlation between class frequency and performance degradation. Specifically, frequent classes ($e.g.,$ Car, Sign, Light) exhibit minimal performance drops, whereas rare classes ($e.g.,$ Bike, Motor) suffer more substantial declines. It is worth exploring DS token design techniques that address the class imbalance in future work.
  
        \begin{figure*}[!t]
            \centering
            \includegraphics[width=0.95\linewidth]{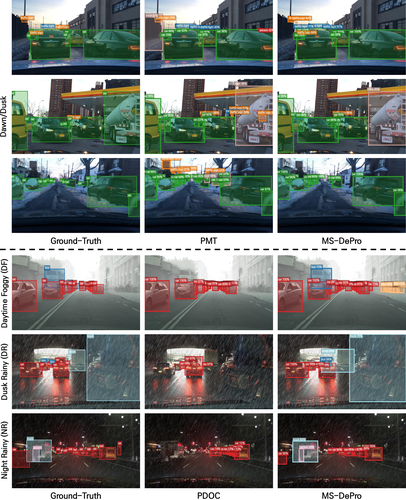}
            \caption{(Top) Detection visualization results in the cross-time domain adaptation. We qualitatively compare MS-DePro with ground-truth and PMT~\cite{belal2024multi}. (Bottom) Detection visualization results in diverse weather domain generalization. We qualitatively compare MS-DePro with ground-truth and PDOC~\cite{li2024prompt}.}
            \label{fig:visualization_DA+DG}
        \end{figure*}
    
        \subsubsection{Visualization Results}
        \label{subsec:visualization}
            \Cref{fig:visualization_DA+DG} illustrates object detection visualizations, showing cross-time domain adaptation in the top 3 rows and diverse weather domain generalization in the bottom 3 rows. In the top 3 rows, PMT~\cite{belal2024attention} exhibits more false positives in specific categories (\textit{e.g.}, traffic sign, person). Conversely, MS-DePro demonstrates remarkable qualitative results. For the extended domain generalization experiments in the bottom 3 rows, PDOC~\cite{li2024prompt} is susceptible to challenges, especially in night and rainy weather (Dusk Rainy and Night Rainy). Our MS-DePro successfully detects objects missed by~\cite{li2024prompt}, demonstrating its generalization across diverse weather. Interestingly, in the Daytime Foggy scene, we find that our method detects the ``bike'' class objects that do not appear to be annotated in the ground-truth.

\section{Conclusion}
\label{sec:conclusion}
    In this paper, we introduce a novel MSDA framework, MS-DePro. Our key idea is the explicit design of domain-invariant representations by leveraging easy-to-access additional modalities—depth map and text. To achieve this, we propose depth-guided localization and multi-modal guided prompt learning. Consequently, MS-DePro achieves SOTA performance on MSDA benchmarks. Furthermore, we extend experiments to the MSDG benchmark, and MS-DePro also demonstrates SOTA results, validating the generalization capabilities of its dedicated components. Despite these impressive results, it also has some limitations. Depth maps are often distorted by motion blur, occlusion, and semi-transparent surfaces, which can degrade model performance by generating inaccurate region proposals and unstable features for the DA token. We leave the task of addressing these challenges to future research to improve overall performance. Ultimately, we believe our work highlights the importance of auxiliary modalities in encoding domain-agnostic features for effective adaptation to the target domain.
    
\section{Acknowledgements}    
    This work was partly supported by the IITP(Institute of Information \& Communications Technology Planning \& Evaluation)-ICAN(ICT Challenge and Advanced Network of HRD)(IITP-2025-RS-2022-00156345, 50\%) and ITRC(Information Technology Research Center)(IITP-2025-RS-2024-004374\allowbreak94, 25\%) grant funded by the Korea government(MSIT). This research was partly supported by Unmanned Vehicles Core Technology Research and Development Program through the National Research Foundation of Korea (NRF) and Unmanned Vehicle Advanced Research Center (UVARC) funded by the Ministry of Science and ICT, the Republic of Korea (NRF-2023M3C1C\allowbreak1A01098408, 25\%).

\clearpage

\appendix

\section{Dataset Details}

    \subsection{Multi-Source Domain Adaptation}
    \label{subsec:appendix_dataset_msda}
        \subsubsection{Cross-time Adaptation}
            In this setting, we train and evaluate on 10 categories in BDD100K~\cite{yu2020bdd100k}: \textit{bike, bus, car, motor, person, rider, traffic light, traffic sign, train, truck}.
            \noindent \textbf{Daytime (D)} consists of 36,728 images, which were captured during the daytime.
            \noindent \textbf{Night (N)} consists of 27,961 images for training.
            \noindent\textbf{Dawn/Dusk (DD)} served as the target domain, selecting 5,027 images for unlabeled domain adaptation and 778 for evaluation.

        \subsubsection{Cross-camera Adaptation}
            In this setting, we train and evaluate only on a \textit{car} class.
            \noindent \textbf{Cityscapes (C)}~\cite{cordts2016cityscapes} is a dataset of high-resolution daytime images from various European cities, from which we selected 2,824 images for training.
            \noindent\textbf{KITTI (K)}~\cite{geiger2012we} is a popular dataset for self-driving research. We selected 6,684 images for training from its \textit{trainval} set, using only those images that include the \textit{car} class.

        \subsubsection{Mixed-domain Adaptation}
            For our experiments, we train and evaluate on 7 categories common to the source and target domain: \textit{person, car, rider, truck, motor, bicycle, bus}.
            \noindent \textbf{Cityscapes (C).} We selected 2,975 images from \textbf{Cityscapes (C)}~\cite{cordts2016cityscapes} for training. For the target domain, we used 36,728 images for adaptation and 5,528 for evaluation
            \noindent \textbf{MS COCO (M)}~\cite{lin2014microsoft} is a large-scale computer vision benchmark. We selected 70,082 training images containing objects from 6 categories, as the \textit{rider} class is not available.
            \noindent \textbf{Synscapes (S)}~\cite{wrenninge2018synscapes} is a synthetic dataset for autonomous driving, simulating various sensor and illumination effects. We selected its entire set of 25,000 images.

    \subsection{Multi-Source Domain Generalization}
    \label{subsec:appendix_dataset_msdg}
        \subsubsection{Diverse Weather Domain Generalization}
            In our experiments, we train and evaluate on 7 categories common to the source and target domain: \textit{bus, bike, car, motor, person, rider, truck}.
            \noindent \textbf{Daytime Sunny (DS)} is an urban-scene detection subset from BDD100K~\cite{yu2020bdd100k}. We selected 19,318 images for training and 8,313 for evaluation from it.     
            \noindent \textbf{Night Clear (NC)}, a BDD100K subset captured at night. We followed the setup of~\cite{vidit2023clip} and selected 7,756 images for training and 27,961 for testing.
            \noindent \textbf{Daytime Foggy (DF)} served as a target evaluation domain with 3,775 images. It combines synthetic data (FoggyCityscapes~\cite{sakaridis2018semantic}) and real images (AdverseWeather~\cite{hassaballah2020vehicle}) to create domain shifts in weather, style, and camera parameters.
            \noindent \textbf{Dusk Rainy (DR)} is a collection of rendered images designed to simulate rainy conditions. It comprises 3,501 images generated from the \textit{Dawn/Dusk} subset of BDD100K.
            
            \noindent \textbf{Night Rainy (NR)}~\cite{wu2021vector} is synthesized from Night Clear. We selected 2,494 images.

        \subsubsection{Real to Artistic Domain Generalization}
            We used real-world datasets Pascal VOC~\cite{Everingham15} and MS COCO~\cite{lin2014microsoft}, alongside art datasets Clipart1k, Comic2k, and Watercolor2k~\cite{inoue2018cross}. We trained the model on 20 classes, using the common categories from Pascal VOC and MS COCO. While Clipart1k shares these 20 classes, Comic2k and Watercolor2k contain only 6 classes (\textit{bicycle, bird, car, cat, dog, person}). Consequently, Clipart1k was evaluated on 20 classes, while Comic2k and Watercolor2k were evaluated on 6 classes.      
            We used the \textbf{Pascal VOC (VOC)}~\cite{Everingham15} dataset. We aggregated 16,551 \textit{trainval} images (VOC2007 / 2012) for training and used the 4,952 images from the VOC2007 test set for in-domain evaluation, following~\cite{inoue2018cross}.
            \noindent \textbf{MS COCO (M)}~\cite{lin2014microsoft} is a large-scale real-world dataset with 80 categories. From its 118,287 training images, we selected 95,279 images containing objects from the 20 categories it shares with Pascal VOC. 
            \noindent \textbf{Clipart1k (CP)}~\cite{inoue2018cross} is an art-domain benchmark sourced from CMPlaces~\cite{castrejon2016learning}, OpenClipart, and Pixabay. We used all 1,000 images (3,165 instances) for out-domain evaluation.
            We selected the 1,000-image test set from \textbf{Comic2k (CM)}~\cite{inoue2018cross}, an art-style dataset sourced from BAM!~\cite{wilber2017bam}, for evaluation.
            \noindent \textbf{Watercolor2k (WC)}~\cite{inoue2018cross} is an art-domain benchmark, primarily of the \textit{bird} and \textit{person} classes. We selected its 1,000-image test set for evaluation.

\bibliographystyle{elsarticle-num} 
\bibliography{main}
\end{document}